# edATLAS: An Efficient Disambiguation Algorithm for Texting in Languages with Abugida Scripts

Sourav Ghosh, Sourabh Vasant Gothe, Chandramouli Sanchi, Barath Raj Kandur Raja
Samsung R&D Institute Bangalore, Karnataka, India 560037
Email: { sourav.ghosh, sourab.gothe, cm.sanchi, barathraj.kr }@samsung.com

*Abstract*—Abugida refers to a phonogram writing system where each syllable is represented using a single consonant or typographic ligature, along with a default vowel or optional diacritic(s) to denote other vowels. However, texting in these languages has some unique challenges in spite of the advent of devices with soft keyboard supporting custom key layouts. The number of characters in these languages is large enough to require characters to be spread over multiple views in the layout. Having to switch between views many times to type a single word hinders the natural thought process. This prevents popular usage of native keyboard layouts. On the other hand, supporting romanized scripts (native words transcribed using Latin characters) with language model based suggestions is also set back by the lack of uniform romanization rules.

To this end, we propose a disambiguation algorithm and showcase its usefulness in two novel mutually non-exclusive input methods for languages natively using the abugida writing system: (a) disambiguation of ambiguous input for abugida scripts, and (b) disambiguation of word variants in romanized scripts. We benchmark these approaches using public datasets, and show an improvement in typing speed by $19.49\%$, $25.13\%$, and $14.89\%$, in Hindi, Bengali, and Thai, respectively, using Ambiguous Input, owing to the human ease of locating keys combined with the efficiency of our inference method. Our Word Variant Disambiguation (WDA) maps valid variants of romanized words, previously treated as Out-of-Vocab, to a vocabulary of $100k$ words with high accuracy, leading to an increase in Error Correction F1 score by $10.03\%$ and Next Word Prediction (NWP) by $62.50\%$ on average.

*Index Terms*—abugida, alphasyllabary, ambiguous input, word variant, multilingual texting, romanization, language modelling, mobile devices, natural language processing, soft keyboard.

## I. Introduction

Keyboard layouts like QWERTY have been ubiquitous and have percolated to soft keyboards in modern smartphones. Yet such layouts are not accommodative for typing in languages with alphasyllabary or abugida scripts, which represent each syllable with a single or conjunct consonant along with optional diacritics denoting vowels [1]. A majority of languages belonging to India and other South Asian countries are influenced by Brahmi [2] and are written using abugida scripts. Typically, each of these languages has a character set of at least 50 distinct unitary characters, apart from the typographic ligatures formed by consonant-consonant conjunctions.

Big data analytics of 3 Samsung Galaxy smartphones, all launched in 2019-2020, reveal that across top-5 Hindi speaking states of India (by native speaker population [3]), the average number of Hindi language users who prefer to type in native layouts exceeds the number of users in the same demography who prefer transliteration layouts [4] by a very large margin, as seen in Figure 2. Even with such a significant user base observed for other abugida script languages, there are multiple pain points for an end-user to text in native scripts [5]. Firstly, singular layouts with regional scripts are not pervasive as it is not feasible to accommodate 50+ native characters in any single keyboard layout. In physical keyboards, usually, each character key is mapped to two characters, one of which can be typed directly and the other with the use of the Shift key. In soft keyboards, a more popular approach is to have multiple views, each with a portion of the character set. Users are expected to use a Switch key which cycles across the views. However, having to switch views multiple times just to type a single word increases the average number of keystrokes required to type as well as proves to be an obstruction to the user's natural thought process.

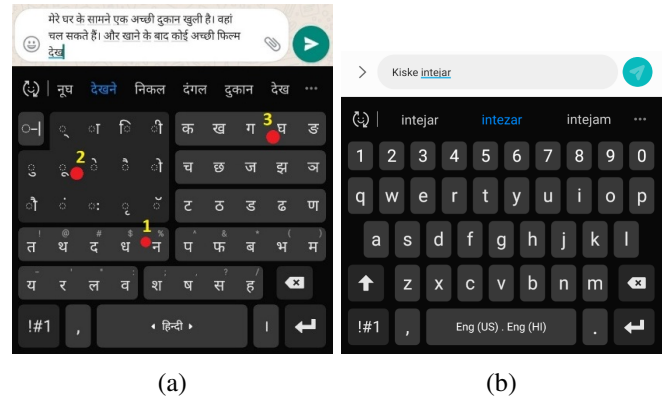

Fig. 1: Application of edATLAS in (a) ambiguous input for abugida scripts, and (b) word variants for romanized scripts

This leads many users to prefer typing in romanized script via more familiar keyboard layouts like QWERTY, which comes with its own challenge. In spite of the popularity of QWERTY, the layout is comprised of Latin characters which cannot accurately represent most syllables of abugida script languages, necessitating a one-to-many approximation which varies from user to user. We refer to these multiple romanizations of the same word as *"Word Variants"*. These further make it difficult for an on-device language model to accurately map a user typed token to the intended native vocabulary word. Such issues also impede the end-user experience with Multilingual Texting and Language Detection [6].

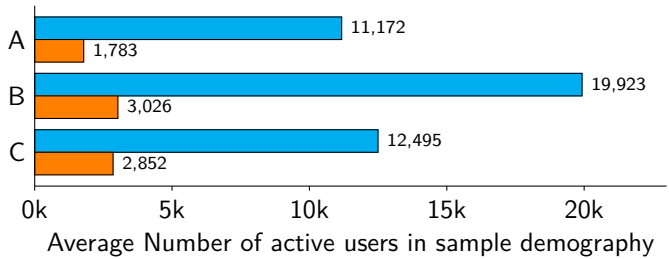

Fig. 2: Usage statistics of Samsung Galaxy smartphones (launched: 2019-2020): Models A, B, C (representative)
■: Native Layout (Hindi input with Devanagari keyboard)
■: Transliteration Layout (Hindi input with QWERTY keyboard)

In the present work, we address these problems by proposing **a novel disambiguation algorithm for two mutually non-exclusive input methods [7] for languages with abugida scripts**. In Sections III and IV, we discuss our proposed algorithm while demonstrating its application in these two input methods. The first application of our proposal is towards the **disambiguation of ambiguous input [8] for native abugida script layouts** that allows a user to interact with a simpler single-view key layout, with a group of related characters represented on a single key. This bypasses the need to explicitly switch between layout views. The user's intended word is inferred by our disambiguation algorithm and the text field is populated in real-time. The second application of our proposal is towards the **disambiguation of word variants in romanized script layouts** that efficiently maps user's preferred variants with their corresponding vocabulary words while maintaining minimal vocabulary size. In both cases, our models complement an existing language model. To the best of our knowledge, no similar solution has been proposed in the literature. We benchmark these approaches using public datasets and Op-Ngram [9] as the base language model. Throughout this work, we use Indic languages and one non-Indic language (Thai) for examples and evaluations; however, the proposals can be directly extended and applied to the entire class of languages with abugida scripts. In our experiments, we maintain a small vocabulary size of $100k$ words as the proposals are targeted primarily for mobile devices with resource-constraints.

## II. Related Work

Asian languages like Chinese, Japanese, and Korean have a large number of ligatures, which has attracted enough research attention to their input systems. However, abugida script languages pose some unique challenges:

1) Defined algorithm exists to decompose any Korean Hangul syllable into a sequence of 2 or 3 Hangul Jamo characters. The set of these characters are comparatively much smaller and can be represented with relative ease in popular layouts, like 2-bulsik [10], which have a similar number of keys as QWERTY. In contrast, abugida scripts have a **large number of unitary ligatures that cannot be decomposed further**.
2) Chinese has an official romanization system: Hanyu Pinyin [11], which drives how a native character is to be represented using the Roman alphabet. However, most abugida languages like Hindi, Bengali, Thai, Telugu, **do not have any such official standards**. Thus, romanization of the same native phoneme varies from person-to-person, and at times, from word-to-word.
3) Input method like Wubizixing or Wubi [12] for Chinese relies on the fact that the native characters are written in a **defined set and order of strokes, which is not applicable** for most, if not all, abugida languages.

In an approach to simplify abugida, proposed by Ding et al. [13], most diacritics are omitted and remaining characters are merged. Dhore et al. [14] present a transliteration method for Hindi and Marathi languages to English, based on the number of diacritics used to form a phoneme and length of named entity. However, these works do not discuss any word recommendation method with intent to aid end-user typing experience. Romanization of Khmer and Burmese languages has been investigated with a statistical approach [15]. But this is not useful in abugida as the mapping between abugida and Latin scripts is complex, where many-to-many alignment between characters is common and it is difficult to statistically model without any heuristics. Our current work addresses this particular issue by capturing a large set of possible word variants which we discuss in section IV. Although there have been attempts to propose alternative input method systems [16] [17] [18], the aforesaid challenges have held them from gaining traction.

Thus, the existing work for large character-set languages leverage character decomposition, standardized transcription, stroke order, statistics, etc. However, the high number of unitary single-Unicode ligatures, lack of standardized romanization rules, and no applicable stroke-system of writing make it challenging to simplify input methods for languages with abugida scripts. Our work differs from the existing by focusing on the phonetic similarity of abugida script characters for both native and romanized layouts, and thereby, proposing a fast-inference, yet low-resource, disambiguation algorithm.

## III. Disambiguation of Ambiguous Input for Native Abugida Scripts

Ambiguous input method [8] relates to a keyboard design with a simplified key layout accommodating less number of keys relative to the typeable characters. This is accomplished by representing related or sequential characters on a single key. The user is provided with the flexibility to tap anywhere on the key containing her desired character, and the input engine predicts the desired character in real-time based on context (preceding words which are already committed in the input field) and vocabulary. This is expected to significantly reduce fat-finger typing errors over conventional layouts. It may be noted that this is dissimilar to approaches like Multi-tap [19] which presents $3 \times 4$ layouts where the user has to explicitly tap a key multiple times to denote the index of intended character, which increases the number of required keystrokes and reduces typing speed. Instead, we take inspiration from predictive text technologies for smaller character set scripts like T9 [20] [21] and XT9 [22], which attempt to match multiple character permutations against a dictionary to filter valid words and suggest the user's intended word. Our proposal limits the maximum keystroke count for a word to its character-length. Furthermore, the grouping of characters is proposed on the basis of linguistic similarities of phonemes as detailed in the following subsections.

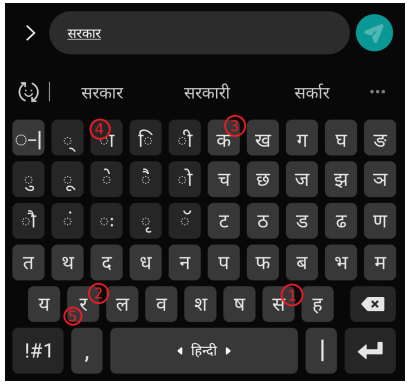

Fig. 3: Typical Conventional Layout (Hindi) (①-⑤ : स + र + क + ा + र ⟹ सरकार )

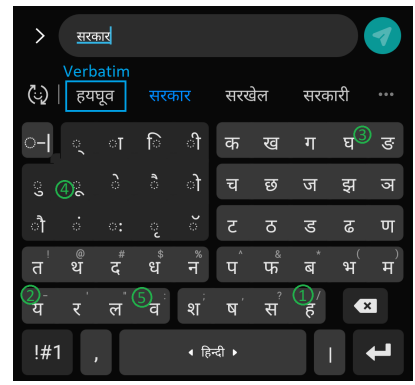

Fig. 4: Ambiguous Input Layout (Hindi) (①-⑤ : शषसह + यरलव + कखगघङ + VOWEL + यरलव ⟹ सरकार )

### A. Layout Design Decisions

While decreasing the number of keys provides ease of use to the user, it leads to an increase in ambiguity. To find a balance in this trade-off, we estimate the predictability of key sequences in different layouts by varying the number of consonants mapped to a key. We do this by computing the number of vocabulary words that get mapped to the same key sequences such that they cannot be predicted with 100% accuracy in a set of maximum 3 suggestions. This analysis was performed with top-2 Indian languages (in terms of the number of native speakers [3]) - Hindi and Bengali, and the official language of Thailand, Siamese or Thai. In Hindi, we observe that even after decreasing the number of keys from over 50 to only 7 consonant keys (containing an average of 4.71 consonants per key) and only 1 vowel key, 89.88% key sequences can be accurately predicted even without any context information. Similar trend is observed for Bengali and Thai as shown in Figure 5. For each language, we determine the optimal layout using the elbow point in the graph.

Established on our observations and logical similarity based on which characters are typically grouped together, in our implementation with Indian languages, we grouped all vowels into one key and the consonant characters based on their phonetic similarities. For example, in the Hindi language written in Devanagari script, all vowels (अ, आ, इ, ई, उ, ऊ, ऋ, ए, ऐ, ओ, औ) are represented on a single key, velar consonants (क, ख, ग, घ, ङ) are presented as one key, palatal ones (च, छ, ज, झ, ञ) as another, and so on. The vowel key stands to also represent the diacritic forms of the vowels. We use the term *Sibling Characters* to denote all characters which are mapped to the same key. Figures 3 and 4 show the sample of a typical conventional layout for a keyboard supporting Hindi in

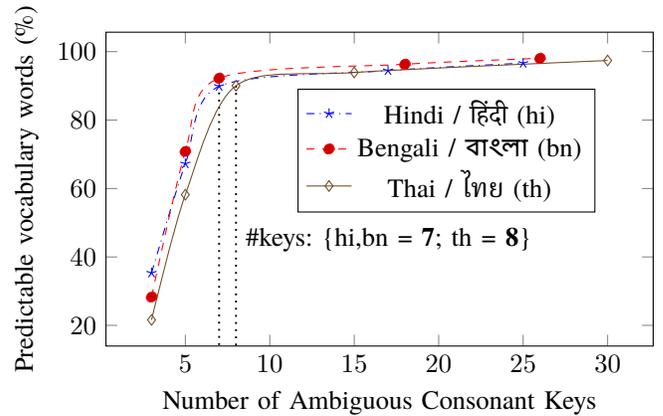

Fig. 5: Vocabulary which can be predicted accurately (top-3 suggestions | empty-context) in different key layouts

Devanagari script and a corresponding ambiguous input layout, where we have retained the relative position of characters in the layout and only combined the keys containing the characters. In the figures, the numbers enclosed in circles denote sample keystroke sequences required to type the Hindi word "सरकार" in both the layouts.

### B. Corpus Description

We prepare training corpus by crawling sources in the public domain such as news articles and blog posts as well as proprietary materials, including linguist-verified user posts in colloquial language, subtitles from movies, volunteer user contributions, etc. For each language, the corpora consist of approximately 50 million distinct sentences. For evaluation set, we use public-domain datasets in Hindi, Bengali and Thai [23] [24] [25], which we describe further in Section III-E.

### C. Preload Model Generation

We collectively refer to the language model components which come out-of-the-box in a smartphone when the language is downloaded as "Preload Model". We define the ambiguous representation, $\overline{w}$, of a word, $w$, to be a minimal variant, which can be computed from $w$ with the application of a set of rules, $\mathbf{R}_l$. The rules are unique for a language $l$ with a given layout, and are designed in such a way that all character sequences, which can be derived by substituting any of the characters in $w$ by one of its Sibling Character, share the same ambiguous representation. Simply put, two words with identical input key sequence also have the same ambiguous representation. Mathematically,

$$\overline{w} = \text{transform}(w, \mathbf{R}_l) \tag{1}$$

Every key is assigned a *Representative Character*. In our implementation, we use the character having the least Unicode value among all Siblings for this purpose. The rules, $\mathbf{R}_l$, simply map each character to its corresponding Representative Character. Table I shows a few examples of word transformations from Hindi and Bengali. The first two examples in each language show different words getting mapped to the same ambiguous representation.

TABLE I: Ambiguous Representation Examples (∗ Different base words mapped to identical ambiguous representation)

| Language | Word | edATLAS Ambig. Repr. | Index: Rule (1 char) |
|---|---|---|---|
| Hindi | घर | कय* | 1: (क\|ख\|ग\|घ\|ङ → क) |
| | कल | कय* | 2: (य\|र\|ल\|व → य) |
| | कर्तव्य | कयअतयअय | 2: (य\|र\|ल\|व → य) |
| | वास्तविक | यअशअतअअक | 2: (Vowel → अ) |
| Bengali | কনে | কতঅअ* | 1; (ক\|খ\|গ\|ঘ\|ঙ → ক) |
| | গদা | কতঅअ* | 2: (ত\|থ\|দ\|ধ\|ন → ত) |
| | হাস্যকর | শঅশঅযকয | 3; (শ\|ষ\|স\|হ → শ) |
| | বিদেশ | পঅতঅশ | 2: (Vowel → অ) |

We use Op-Ngram [9] as the underlying language model (LM) for suggestions, which has a vocabulary of top-$100k$ words for each language, stored in a Marisa Trie [26], ranked according to their statistical probability in the training corpus. This trie is referred to as VocabTrie. We choose Op-Ngram due to its low-resource requirements, which is ideal for on-device inferencing on mobile devices. Next, we generate a corresponding ambiguous vocabulary where each vocabulary word is to be substituted by its ambiguous representation. This indexed ambiguous vocabulary is then used to generate a parallel Marisa Trie, termed AmbigTrie. The AmbigTrie words retain the same ranks, or Marisa Trie indices, as those in VocabTrie. Our average AmbigTrie size per language with top-$100k$ words is $602$ KB compared to the average source VocabTrie size of $834$ KB.

*D. Inference*

When user types using ambiguous input layout, token entered by the user is fed to the transform() function along with the same set of rules $\mathbf{R}_l$ to generate its ambiguous representation. This step is integrated with IME [7] by mapping each key with its Representative Character. The ambiguous representation $\overline{w}$, thus generated, is then used to perform a prefix search for word completion or correction candidates in AmbigTrie (without Bloom filter). Using each result, potential vocabulary words are fetched from VocabTrie with the help of common word indices between the tries. These vocabulary words are then sorted based on a confidence score determined by the word length, preceding context, and language model. Vocabulary word with the highest net score is used as partial input in the text field and as an Auto-Completion candidate.

*E. Evaluation*

To evaluate our proposal, we ask 200 Beta-trial users to type a set of 50 pre-defined sentences in both conventional and proposed ambiguous input layouts using identical Samsung Galaxy S10 smartphones. For evaluation, we use random sample from "Code-Mixed (Hindi-English) Dataset" [23], "Indian Language Part-of-Speech Tagset: Bengali" [24], and "ThaiGov V2 Corpus" [25], containing news articles for Hindi, Bengali and Thai. Sentences are sampled such that each example has 15 to 35 characters (both inclusive), without considering punctuation marks. To minimize any potential bias due to the relative position of characters on screen, we use the same underlying layout in both cases. The only difference is that in the former, all characters are represented separately and a switch key is used to toggle between available views; while in the latter, the keys for Sibling Characters are joined together and presented as one and the view switch key is absent or non-functional.

We use two metrics to benchmark the quality of LM as perceived by an end-user, namely Keystroke Saving Ratio (KSR) and Next Word Prediction (NWP) with a sample test set of 500 sentences [23], [24], [25]. The KSR metric [27] indicates the number of keystrokes saved due to effective suggestions provided by the engine. It is defined as:

$$\text{KSR} = \frac{n_c - n_k}{n_c} \times 100\% \tag{2}$$

where $n_c$ denotes the number of actual characters in the test set and $n_k$ denotes the number of keystrokes that were needed to produce the test set with a soft keyboard. The NWP metric [9] refers to the probability that given a sequence of the first $(k-1)$ words from a sentence in test data, the next word in that sentence, $w_k$, will appear among the word prediction candidates. Mathematically,

$$\text{NWP} = \frac{\sum_{\substack{s \in \mathbf{S}, \\ k \in [1, n_w^s]}} \left| \left\{ w_k^s \middle| w_k^s \in \mathbf{W}_k^s \right\} \right|}{\sum_{s \in \mathbf{S}} n_w^s} \times 100\% \tag{3}$$

where $\mathbf{S}$ refers to the set of all sentences in the test set, $w_k^s$ refers to actual test set word at $k^{\text{th}}$ position in sentence $s$, $\mathbf{W}_k^s$ denote the set of suggestions provided by the inference engine as candidates for word $w_k^s$, and $n_w^s$ indicates the total number of words in sentence $s$ of test set. In our evaluation, we restrict the size of $\mathbf{W}_k^s$ to only a maximum of top-3 word suggestion candidates.

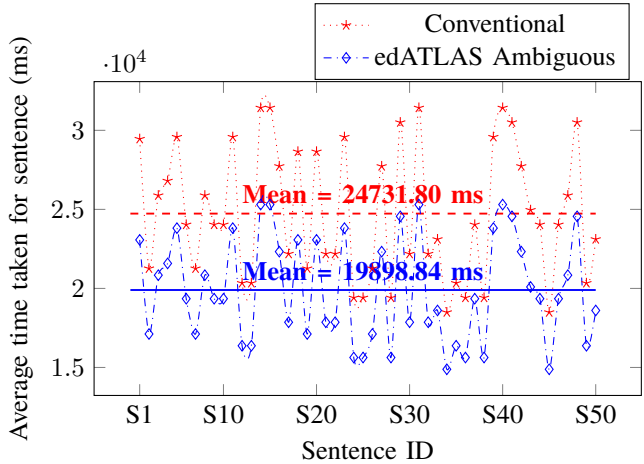

Fig. 6: Average time taken to type each sentence in test set (Hindi)

Mean over sentences: ⋆ 24.7 seconds → ⋄ 19.9 seconds

TABLE II: Impact of edATLAS Ambiguous Input on Typing Speed (Base LM: Op-Ngram [9])

| Language | Average time taken to type one character | | |
|---|---|---|---|
| | Samsung Keyboard Conventional (ms) | Samsung Keyboard + edATLAS Ambiguous (ms) | Improvement (%) |
| Hindi | 924.32 | **744.18** | 19.49 |
| Bengali | 940.33 | **704.02** | 25.13 |
| Thai | 1232.68 | **1049.08** | 14.89 |

TABLE III: Impact of edATLAS Ambiguous Input on Next Word Prediction (NWP) and Keystroke Saving Ratio (KSR) (∗ Minor deterioration in KSR is *normal and expected* as we are introducing ambiguity in key determination)

| Language, Dataset | NWP | | | KSR | | |
|---|---|---|---|---|---|---|
| | Op-Ngram + Conv. | Op-Ngram + edATLAS Ambig. | ↓ (%) | Op-Ngram + Conv. | Op-Ngram + edATLAS Ambig. | ↓ (%) |
| Hindi, [23] | 30.31 | 30.31 | 0 | 50.12 | 50.03 | * 0.18 |
| Bengali, [24] | 12.27 | 12.27 | 0 | 49.99 | 48.07 | * 3.84 |
| Thai, [25] | 11.53 | 11.53 | 0 | 47.39 | 47.06 | * 0.70 |

Figure 6 shows the average time taken to type each sentence in the set of 50 sentences for Hindi. Table II demonstrates that the average time taken to type a character is drastically improved with edATLAS ambiguous input layout. We further evaluate average addition to inference time as only 1.4 ms.

Now, the expectation of better accuracy with any proposal is quite justified in general. However, in the case of Ambiguous Input, the number of characters per key are more than one, unlike conventional keyboard layouts. As the user can type anywhere in the key, the intended character needs to be resolved by the engine at runtime. As a result, we are introducing a degree of ambiguity in character determination. When one-to-one mapping exists as in conventional non-ambiguous layouts, accuracy can be almost 100% for an ideal typist who does not make typing errors. However, even when this ideal typist taps an ambiguous key, there is more than one possibility for the desired character. Being a probabilistic model, there exists a margin of error during the first one or two characters of a word, especially when the already typed context is short. The quality of underlying LM can improve the accuracy but practically it can never reach 100%. Due to this, minor deterioration of accuracy is difficult to get rid of completely.

In spite of this, Table III shows that the deterioration of the Keystroke Saving Ratio (KSR) is negligible and that there is no impact on the NWP metric. The zero impact on NWP is because it is dependent only on the preceding context tokens. Moreover, as we can observe from Table II, the minor KSR drop is more than made up for by the drastic improvement in the average time taken for an end-user to type a character. This significant improvement in typing speed with only a minor impact on KSR reaffirms the optimality of our chosen layout in addition to the elbows in Figure 5.

## IV. Disambiguation of Word Variants in Romanized Scripts

In this section, we discuss our solution to a crucial problem faced while texting in romanized script for those languages where no predefined or one singular set of widely accepted romanization rules is available. This is especially relevant to languages with abugida scripts as the ISO basic Latin alphabet [28] contains 26 characters with only 5 vowels, which is not sufficient to represent all vowels present in most abugida scripts. In the absence of pre-defined rules, each user develops her own empirical logic to romanize native words. These rules may not only vary widely between two people but often vary when the same user is typing multiple words.

Let us represent romanizations of sample sentences as "$\nu_1$ <$\varrho_1^1$ | $\varrho_1^2$ | ..>, $\nu_2$ <$\varrho_2^1$ | $\varrho_2^2$ | ..>, ...", where $\nu_i$ is the $i^{th}$ word in native script and $\varrho_i^j$ is the $j^{th}$ alternative for its romanization. Then, romanization of a typical Hindi sentence can be *"काफी <Kaafi | Kafi | Kafee> देर <der> से <se> इंतज़ार <intezaar | intejaar | intejar | intezar> कर <kar> रहे <rahe | rahein> हैं <hain | hai>, कहा <kaha | kahaa> हो <ho>?"* Again, a Thai sentence may be romanized as: *"ฉัน <Chan | Chaan>, ชอบ <chxb | chhp>, ทาน <than | tan>, ผัด <Pad | Phad | Phat>, ไทย <Thai>"*.

Thus, there is ample ambiguity for a language model to recognize so many word variants and interpret it as a known vocabulary word. Firstly, each variant not present in the limited vocabulary often gets interpreted as a typo and may get auto-corrected to a nearby spelled word, which may be different from the user's intended word. Secondly, even if the word gets entered by the user successfully, it cannot be mapped to a vocabulary word and gets treated as an unknown Out-Of-Vocabulary (OOV) token, leading to poor quality of subsequent Next Word Prediction (NWP). Thirdly, such variant words entered by the user get learned by the system only in the context where they had been entered, and their usage is treated separately from the In-Vocabulary variant. Thus, they cannot be suggested back to the user in future relevant scenarios. Our proposal addresses these problems by storing a parallel ambiguous representation for each vocabulary word, which can capture multiple possible word variants.

## A. Preload Model Generation

As in Section III, we use a transform() function and a set of rules, $\mathbf{R}_l$, to map each word to a variant representation. However, here, ruleset $\mathbf{R}_l$ is determined on the basis of linguistic inputs regarding the interchangeability of character sequences for a language $l$. Table IV shows a complete set of equivalence relations, $P \simeq Q$, that we use in our models for 3 languages: Hinglish (romanized Hindi), Benglish (romanized Bengali), and Tenglish (romanized Telugu). These equivalence relations are used to generate the ruleset, $\mathbf{R}_l$, by mapping both the elements in the columns, labeled '$P$' and lexicographically preceding '$Q$', to ambiguous representation denoted by toUpperCase($Q$). In an ideal scenario, *all* words which are variants of each other get assigned the same base representation, and non-variant words differ in their variant representations.

TABLE IV: Equivalence Relations : Basis of Variant Rules
$(P > Q) \wedge (P \simeq Q) \implies (P|Q \to \text{toUpperCase}(Q)) \in \mathbf{R}_l$
'<' and '>' denote start and end of a word, respectively.

| Hinglish | | Benglish | | Marathinglish | | Tenglish | |
|---|---|---|---|---|---|---|---|
| P | Q | P | Q | P | Q | P | Q |
| aa | a | aa | a | aa | a | <ye | <e |
| ain> | ai> | ai | oi | cc | c | aa | a |
| cc | c | au | ou | ee | i | au | ou |
| chch | ch | bh | v | hh | h | bh | b |
| ee | i | cc | c | ii | i | cc | c |
| ein> | e> | ee | i | oo | u | chch | ch |
| hh | h | hh | h | ph | f | dh | d |
| ii | i | ii | i | uu | u | ee | i |
| oo | u | ny | nn | w | v | gh | g |
| ph | f | oo | u | z | j | hh | h |
| sh | s | ph | f | | | ii | i |
| uu | u | sh | s | | | kh | k |
| w | v | uu | u | | | ly> | li> |
| z | j | ye | e | | | oo | u |
| | | z | j | | | ov | ou |
| | | | | | | sh | s |
| | | | | | | th | t |
| | | | | | | uu | u |
| | | | | | | w | v |
| | | | | | | z | j |

The ambiguous base representations are used to generate a variant vocabulary Marisa Trie, namely BaseTrie, where words share the same ranks as their VocabTrie counterparts. Table V depicts some sample transformations for Hinglish, Benglish, and Tenglish. For these languages with an average VocabTrie size of 731 KB containing top-100$k$ words, the average BaseTrie size is as low as 562 KB.

## B. Composing Word Identification

When the user starts typing a token (*composing word*), the partial word is converted to its variant representation, and the same is used to perform a prefix lookup in both the general and base vocabularies to fetch completion and correction candidates. As earlier, suggestions from BaseTrie are substituted with corresponding VocabTrie words using shared Marisa Trie indices. Based on context words, edit distance, confidence scores, and origin of suggestion, these candidates are sorted and used to populate suggestions. If the input token is not found in VocabTrie but is an exact match (or a very close match in terms of edit distance) to a candidate derived from BaseTrie, then the same is offered as an Auto-Correction candidate for user token. The user still gets the choice to override Auto Correction and enter the verbatim token. Variant suggestions, with comparatively higher edit distances, are offered as Correction candidates but not as Auto-Correction to respect user preferences.

TABLE V: Variant Representation Examples

| Language | Word | edATLAS Variant Representation |
|---|---|---|
| Hinglish | pradhanmantree | prAdhAnmAntrI |
| | pradhaanmantri | prAdhAnmAntrI |
| | rashtriya | rAStrIyA |
| | hain | hAI |
| Benglish | dhormiyo | dhOrmIyO |
| | dharmiya | dhOrmIyO |
| | debota | debOtA |
| | debotaa | debOtA |
| Tenglish | ekkada | EkkADA |
| | yekkada | EkkADA |
| | apoorvaa | ApUrVA |
| | apurva | ApUrVA |

## C. Next Word Prediction with Context containing Variant(s)

Context user input, i.e., leading words that have already been entered, is used by the language model to determine the likelihood of next word suggestion candidates. Typically, when a language model encounters Out-of-Vocabulary (OOV) words in the context, it treats them as generic unknown tokens. This leads to irrelevant suggestions and affects the texting experience. To address this, we substitute all OOV words in context, wherever possible, with a vocabulary word derived by transforming the OOV token to its base form, fetching all possible words using Marisa Trie indices, and then selecting the one with the highest confidence score with the remaining context. In case no viable substitute is available for an OOV token, it is retained as it is. Figure 7 presents a flowchart for disambiguating word variants in context.

## D. Personalisation of User Language Model

Our approach inherently supports two different scenarios of learning variant-usage-pattern from end-user. Firstly, if the user prefers to type a different variant of a word than what is present in VocabTrie, our language model receives the user's preferred token as a composing token and learns it when the user performs a commit action viz. entering space or period mark, sending a message, etc. In the future, this variant is suggested by the language model instead of the VocabTrie variant.

TABLE VI: Impact of edATLAS Word Variant Disambiguation (WVD) on Keystroke Saving Ratio (KSR), Next Word Prediction (NWP), and Error Correction (EC): Precision, Recall, F1 (Before: Op-Ngram; After: Op-Ngram + edATLAS WDA)

| Language | KSR | | | NWP | | | EC Precision | | | EC Recall | | | EC F1 Score | | |
|---|---|---|---|---|---|---|---|---|---|---|---|---|---|---|---|
| | Before | After | ↑ (%) | Before | After | ↑ (%) | Before | After | ↑ (%) | Before | After | ↑ (%) | Before | After | ↑ (%) |
| Hinglish | 27.44 | **33.67** | 22.70 | 16.67 | **28.40** | 70.37 | 0.57 | **0.64** | 12.28 | 0.29 | **0.30** | 3.45 | 0.38 | **0.41** | 7.89 |
| Benglish | 32.15 | **38.94** | 21.12 | 17.93 | **31.16** | 73.79 | 0.53 | **0.66** | 24.53 | 0.32 | **0.35** | 9.38 | 0.40 | **0.46** | 15.00 |
| Marathinglish | 40.48 | **47.57** | 17.51 | 14.55 | **23.49** | 61.44 | 0.48 | **0.52** | 8.33 | 0.43 | **0.44** | 2.33 | 0.45 | **0.48** | 6.67 |
| Tenglish | 26.89 | **27.84** | 3.53 | 9.27 | **13.37** | 44.23 | 0.40 | **0.43** | 7.50 | 0.38 | **0.41** | 7.89 | 0.39 | **0.42** | 7.69 |
| Romanized Thai | 23.28 | **28.39** | 21.95 | 11.26 | **18.32** | 62.70 | 0.39 | **0.45** | 15.38 | 0.26 | **0.28** | 7.69 | 0.31 | **0.35** | 12.90 |

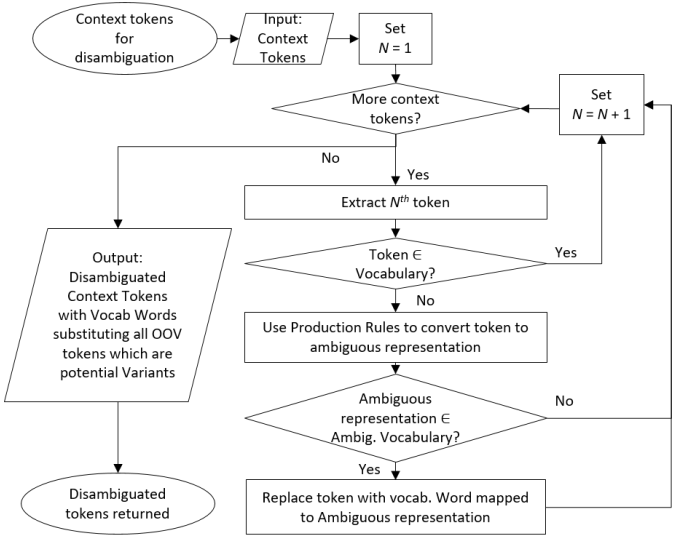

Fig. 7: Disambiguation of OOV Word Variants in Context (words preceding current Composing Token)

Secondly, if the user uses different variants for the same word in multiple scenarios, in all cases, the contexts are processed to substitute OOV variants to identical base representations as in BaseTrie with the application of rules in $\mathbf{R}_l$, and then by corresponding VocabTrie variants having highest language model score. Thus, context information is learned irrespective of which variant the user uses. These aggregate and generate, and continuously update, a mutable language model which is personalized to the end-user. We refer to this as *Variant Agnostic Learning*.

Furthermore, the entire process is performed completely offline, which ensures user data privacy.

### E. Evaluation

To evaluate our proposals for Word Variant Disambiguation, we evaluate the impact on four Indic languages in their romanized scripts: Hinglish (romanized Hindi), Benglish (romanized Bengali), Marathinglish (romanized Marathi), Tenglish (romanized Telugu), and on one non-Indic language, romanized Thai. Our evaluation dataset primarily consists of tweets in Hindi-English [29], Bengali-English [30], Marathi-English, Telugu-English, and Thai [31], filtered by hashtags of trending stories, which is further cleaned by removing English phrases from code-mixed content. For some samples, tweets are transliterated from their native script. This is augmented by literary resources and news articles, romanized independently by members in a team of linguists, each of whom demonstrates different romanization preferences, representative of text typed by a majority of users.

As in Section III, we use Op-Ngram [9] as base LM. The KSR and NWP metrics (Section III-E) are used to compare the improvement due to proposals described in Sections IV-C and IV-D. This evaluation is performed using an Android-based tester application, which simulates the typing of each character in the input test set, and selects the intended word from suggestions, if and when it appears. As observed in Table VI, there is a significant improvement in KSR and NWP which directly translates to fewer keystrokes required to be typed by end-user and better effective typing speed.

Table VI further proves the efficacy of edATLAS induced enhanced Composing Word Identification (Section IV-B) in improving Error Correction. This is partly due to False Corrections being avoided when a valid but Out-of-Vocab (OOV) word variant is entered by the user, and partly due to the enhanced ratio of True Positives attained by mapping user variants and close typos to desired In-Vocabulary token when other parameters like Edit Distance and LM confidence are indecisive. Average addition to inference time is computed to be 1.12 ms.

### V. CONCLUSION

We present solutions to address the problems faced while typing in abugida script languages. For texting in native abugida scripts, we have proposed an ambiguous input method, which is very efficient in terms of typing speed, resource size and inference time. With $100k$ words in VocabTrie of average size 834 KB, the resultant AmbigTrie occupies 602 KB (avg.). However, the typing speed is improved by $19.84\%$ with only a minor $1.57\%$ drop in KSR (avg.). For texting in romanized scripts, we propose Word Variant Disambiguation (WVD) which addresses the crucial problem of managing multiple acceptable word variants arising out of non-standard romanization. Again, with only 562 KB of BaseTrie for underlying VocabTrie of 731 KB (avg.), our approach improves Keystroke Saving Ratio (KSR) by as much as $22.70\%$ in Hinglish, and demonstrates a $62.50\%$ improvement in Next Word Prediction (NWP) with $10.03\%$ increase in Error Correction F1 score (avg.).

During Apr-Sep 2020, edATLAS has shown a direct correlation to a $26\%$ increase in keyboard Monthly Average Users (MAU) and over $30\%$ improvement in Backkey Ratio (ratio of backspaces with respect to all keystrokes) attributed to less typing errors by end-users. We are currently exploring the usefulness of edATLAS WVD in natively Roman-script languages like English to process *baby language*, referring to the usage of terms like "cuuteee" and "soooo much", popular in informal communications among millennials and *Gen Z* users.